\documentclass[10pt,twocolumn,letterpaper]{article}

\usepackage[pagenumbers]{cvpr} 

\usepackage[dvipsnames]{xcolor}
\usepackage{graphicx}

\usepackage{amsmath,amssymb,stmaryrd}  

\definecolor{cvprblue}{rgb}{0.21,0.49,0.74}
\usepackage[pagebackref,breaklinks,colorlinks,citecolor=cvprblue]{hyperref}
\usepackage[dvipsnames]{xcolor}
\usepackage{algorithm}
\usepackage{algpseudocode}

\def\paperID{******} 
\def\confName{CVPR}
\def\confYear{2024}

\title{Hierarchical NeuroSymbolic Approach for Comprehensive and Explainable Action Quality Assessment}

\author{Lauren Okamoto\\
Princeton University\\
{\tt\small lokamoto@princeton.edu}
\and
Paritosh Parmar\\
IHPC, A*STAR, Singapore\\
{\tt\small paritosh.parmar@alumni.unlv.edu}
}

\begin{document}
\maketitle

\begin{abstract}
Action quality assessment (AQA) applies computer vision to quantitatively assess the performance or execution of a human action. Current AQA approaches are end-to-end neural models, which lack transparency and tend to be biased because they are trained on subjective human judgements as ground-truth. To address these issues, we introduce a neuro-symbolic paradigm for AQA, which uses neural networks to abstract interpretable symbols from video data and makes quality assessments by applying rules to those symbols. We take diving as the case study. We found that domain experts prefer our system and find it more informative than purely neural approaches to AQA in diving. Our system also achieves state-of-the-art action recognition and temporal segmentation, and automatically generates a detailed report that breaks the dive down into its elements and provides objective scoring with visual evidence. As verified by a group of domain experts, this report may be used to assist judges in scoring, help train judges, and provide feedback to divers. Annotated training data and code: \url{https://github.com/laurenok24/NSAQA}.
\end{abstract}  
\section{Introduction}
\label{sec:introduction}
Analyzing people’s movement and actions and providing feedback has applications ranging from physical rehabilitation \cite{parmar2016measuring, capecci2019kimore, sardari2020vi} to sports coaching and scoring athletic performance \cite{xu2019learning, li2018scoringnet, liu2020fsd} to surgery skills assessment \cite{doughty2018s, doughty2019pros, liu2021towards}. With the \textit{2024 Paris Olympics} just around the corner, there is heightened interest in \textbf{action quality assessment} (AQA) in the domain of sports such as diving, gymnastics, and figure skating. These automated systems are generally deep learning-based systems. Such neural approaches analyze videos of sports performance and output a score \textit{quantifying how well} the athlete performed. These neural models are trained using scores given out by human judges during past competitions.

\begin{figure}
\small
    \centering
    \captionsetup{type=figure}
    \includegraphics[width=\columnwidth]{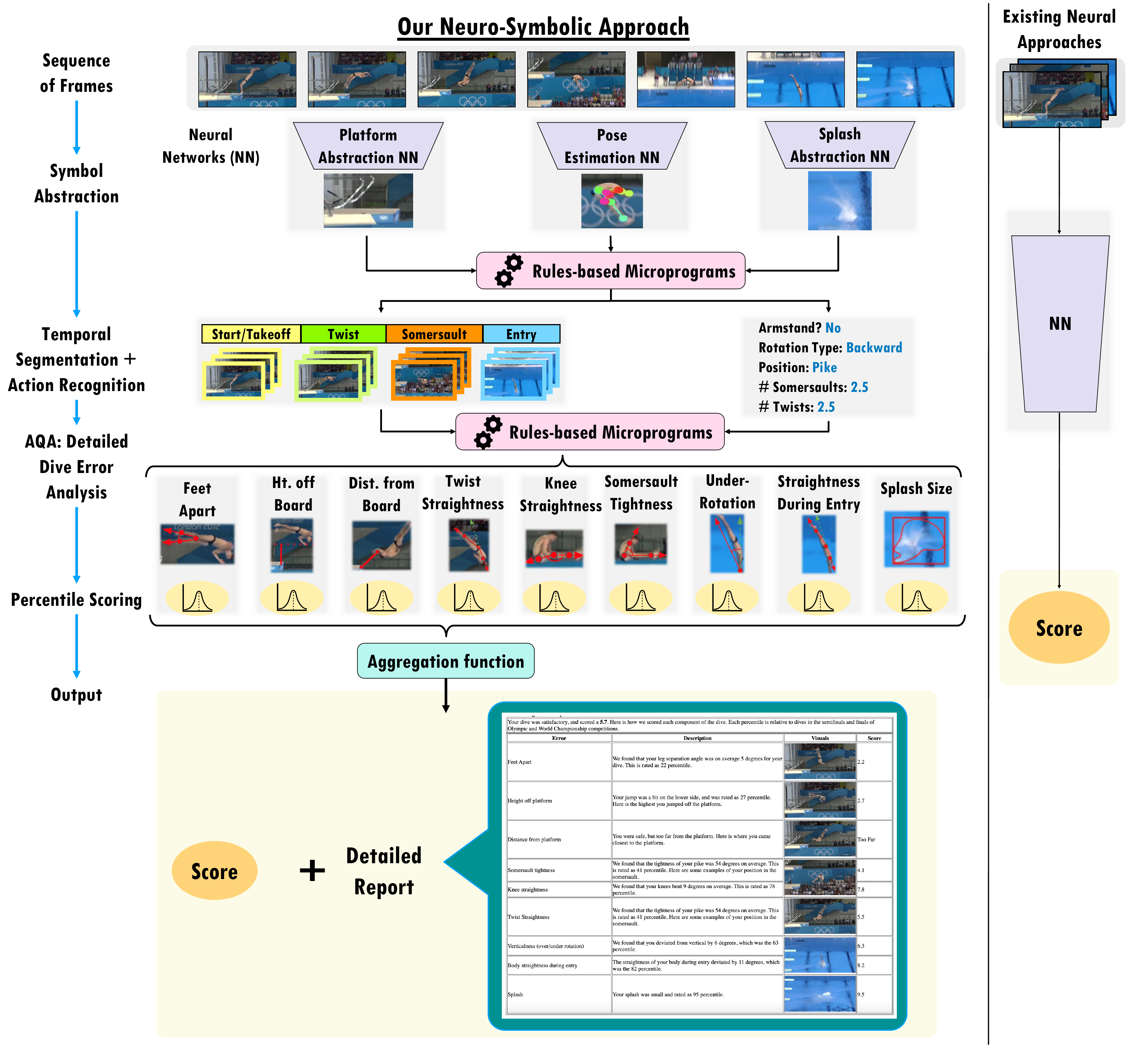}
    \caption{\textbf{Neuro-Symbolic Action Quality Assessment (NS-AQA) vs Neural AQA.} Our NS-AQA approach (Left) employs neural networks to extract crucial symbolic information, such as platform location, framewise pose estimation, \& splash detection. These symbols furnish objective data utilized for rules-based fine-grained action recognition, temporal segmentation, \& detailed error analysis. The outcome is an objective score \& a comprehensive visio-linguistic report, complete with supporting visual evidence, generated programmatically. This is much more valuable than existing AQA approaches (Right) that can only predict a single score (potentially biased) without any accompanying explanation. \textit{Please zoom in; full-size version in supplementary.}}
   \label{fig:teaser}
\end{figure}

The issue is that human judging itself is subjective and prone to bias. The bias may take two forms: bias for/against individual athletes; and bias in how different elements of a performative action are weighted in the overall score. An example of the latter in diving is bias towards the last phase of a dive---entry into the water and the amount of splash created. Because of this, the scores given out by the judges may not take into account all the factors/elements of the athletes’ performance. This problem is further aggravated by the fact that performance scores from human judges are never supported by an actual breakdown of the score or an analysis of what athletes did well or not. This `black box' kind of scoring cannot guarantee that: 1) all the \textit{factors/elements/efforts from athletes’ performance are taken into account} when scoring; \& 2) the \textit{performance has been objectively scored}.  This leads to a lack of trust, fairness, and explainability in the scoring. These issues are compounded when current AQA approaches, which are fully neural models \cite{lei2019survey, doughty2017s, pan2019action, xu2019learning, li2019manipulation, li2018end, sardari2020vi, dadashzadeh2024pecop, zhong2024dancemvp}, are trained using scores given by human judges.  

To mitigate this set of problems, we propose a novel \textit{neuro-symbolic} approach to fine-grained analysis of human movements \& actions. We consider the popular and challenging sport of diving as the case study. Our neuro-symbolic approach combines the strengths of \textit{deep neural models} and \textit{rules-based AI}. Our hierarchical neuro-symbolic approach (see \autoref{fig:teaser}) for fine-grained action analysis first \textit{deconstructs} the whole performance and its surrounding environment/scene into \textit{interpretable} "symbols" such as the platform, athletes’ body poses, water splash, etc. \textit{using neural models}. Then, using a \textit{rules-based} approach \textit{constructs} a \textit{hierarchy of representations}---starting from identifying dive type/class and temporal segments of the dive phases to finally, fine-grained and detailed action quality assessment representation and score. Our system \textit{programmatically} generates \textit{detailed performance analysis reports} that not only check for all the performance errors, but also precisely compute their magnitudes/severities, which is hard to do even for highly-trained eyes, especially, when the actions/movements are high-speed as in case of diving. What is more, our system can sift through the entire performance and collect all the \textit{visual evidence} to support the error detection. With our \textit{transparent} and \textit{objective} approach, performers can see that all their efforts have been rewarded, and bias in judging can be minimized. This is in stark contrast to current AQA approaches, or even human judges, which just give out a \textit{single score without any explanation} of its breakdown.

The contributions of our work can be summarized as follows: 
\begin{itemize}[noitemsep, topsep=0pt]
    \item We propose a Neuro-Symbolic paradigm for AQA that brings remarkable accountability, transparency, and trustworthiness to AQA and Sports Performance Judging. We validated our approach on competitive platform diving.
    \item Our system achieves state-of-the-art (SOTA) performance on fine-grained action recognition and temporal segmentation of dives.
    \item Our system programmatically generates a highly detailed report of the performance, including angular measurements as precise as 1 degree, \& retrieving relevant images \& video clips for the  supporting visual evidence. This report may be used for various purposes: (a) as a potential aid to judges as they score the dive; (b) as a tool to teach judges how to score; (c) to settle disagreements between judges; (d) to encourage safety by penalizing dangerous actions (such as getting too close to hitting the platform); \& (e) as a tool with an intuitive UI for coaches \& divers to detect, quantify, \& visualize performance errors. 
    \item Our system generates statistically-grounded scores that compare a diver’s performance relative to that of their peers. Generated scores are equally useful to all levels of diving competitions (from recreational to Olympics).
    \item We precisely annotate a large number of frames with splash, platform/springboard, and diver segmentation masks to be used for object detection training. We will opensource all the training data and models to help researchers.
\end{itemize}

Our neuro-symbolic approach to AQA can be extended to not only other sports, such as figure skating and gymnastics, but also to other precise complex actions, such as surgery.  In each case, domain expertise is necessary to determine what symbols need to be abstracted and to create suitable rules for analysing and assessing of the action based on those symbols.  We hope that further applications of our NS-AQA system can help reduce bias and objectively grade all sorts of skilled human actions.
\section{Related Work}
Existing approaches treat AQA as a regression problem. Thus far, efforts have focused on predicting action quality scores as closely as possible to those given by human judges. This is achieved through the use of better features \cite{ltsoe}, multitask learning \cite{mtlaqa}, score distribution learning \cite{usdl}, more data \cite{aqa7}, or improved regression techniques \cite{yu2021group, jain2020action}. However, the scores given by human judges, which serve as the `gold standard,' are inherently biased. For instance, in diving, the last stage of the dive or the splash appears to have the most impact on judges, potentially neglecting other crucial factors of the athlete's performance. Therefore, at best, these neural approaches may assess in a biased manner. 

Moreover, existing approaches being purely neural are 'black box' in nature---how they arrive at decisions \& which factors of the performance are considered, as well as how each factor is scored, remain unknown. This lack of transparency results in a dearth of explainability, fairness, \& trustworthiness, making it uncertain whether all the efforts from performers are genuinely rewarded. Ideally, athletes who have sacrificed a significant part of their lives should receive recognition for all their efforts. In contrast, we propose a new paradigm to AQA---a Neuro-Symbolic approach. Our approach leverages modern deep neural networks to extract useful information from performance videos into symbols. It then applies logic to these symbols to make decisions and evaluate performance without relying on biased ground-truths, but rather utilizes domain knowledge for objective scoring. Our system conducts a much more comprehensive and in-depth evaluation, programmatically generating an extremely detailed performance evaluation report. This makes the evaluation process transparent, fair, explainable, and trustworthy. A concurrent work \cite{zhang2024narrative} also recognizes the value of detailed action evaluation; however, it relies on large specialized datasets and, being a purely neural approach, still suffers from previously mentioned issues. Additionally, our reports are more comprehensive and detailed, complete with visual evidence.

Neurosymbolic AI which combines the power of neural models and symbolic systems has been gaining increasing interest \cite{Garcez2012NeuralsymbolicLS, yi2018neural, sun2022neurosymbolic, liang2023logic}, but has not been explored in AQA setting. In this work, we develop a hierarchical neurosymbolic approach to AQA.  
\section{Hierarchical Neuro-Symbolic Approach}
Our neuro-symbolic action analyzer consists of two parts: 1) \textbf{Neural Action-Context Parser}; 2) \textbf{Rules-based Action Analyzer}. The Neural Action-Context Parser \textbf{deconstructs} or decomposes the action or the performance video, extracting meaningful information such as the \textit{pose of the diver} in terms of position of joints, \textit{diving platform segmentation}, and the \textit{splash} created when the diver enters the water. This extracted information forms our \textbf{symbols}, which are then passed to the Rules-based Action Analyzer. The Rules-based Action Analyzer scrutinizes actions through structured and interpretable symbolic reasoning. Note that, our contribution does not lie in proposing merely a two-stage approach, but in integrating neural models, which excel at feature extraction; and symbolic systems, which provide transparent and objective analytics. We will now explain each of these parts in detail.

\begin{table}[]
\small
\centering
\resizebox{\columnwidth}{!}{%
\begin{tabular}{@{} p{0.9\columnwidth} c @{}}
\toprule
\textbf{Platform.} The location of the platform, especially the position of its edge facing the pool, is crucial to determine when the diver leaves the platform, thus starting their dive.  The platform location is also important to assess how close the diver comes to its edge, which is relevant to scoring.   &  
\raisebox{-0.07\textwidth}{\includegraphics[width=0.09\textwidth, height=0.08\textwidth]{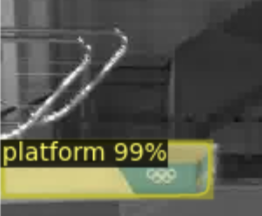}}
 \\ \midrule
\textbf{Splash} at entry into the pool is a conspicuous visual feature of a dive. The size of the splash is an important element in traditional scoring of dives.  A large splash mars the end of a dive and also likely indicates a flaw in form at water entry. &  
 \raisebox{-0.06\textwidth}{\includegraphics[width=0.09\textwidth, height=0.08\textwidth]{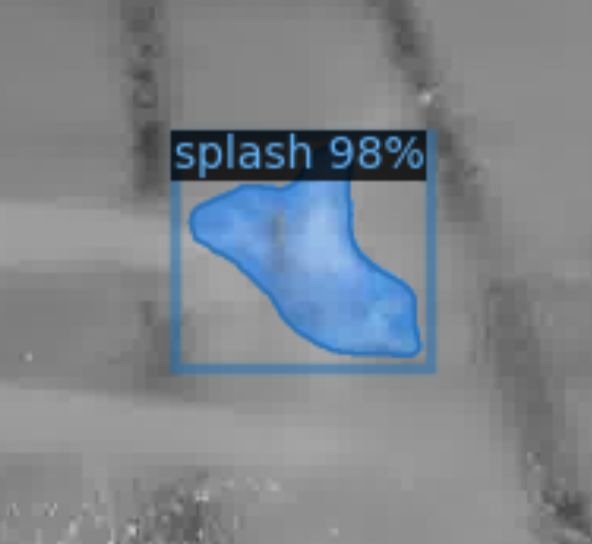}} 
 \\ \midrule
\textbf{Pose} of the diver in video frames is crucial for dive assessment. We gather 2D pose data, detailing body part locations like the head, thorax, pelvis, etc. Analyzing this data frame by frame allows identification of sub-actions (e.g., somersaults, twists) and quality assessment. Detailed pose data enables objective metrics like water entry angle.&  
 \raisebox{-0.07\textwidth}{\includegraphics[width=0.09\textwidth, height=0.09\textwidth]{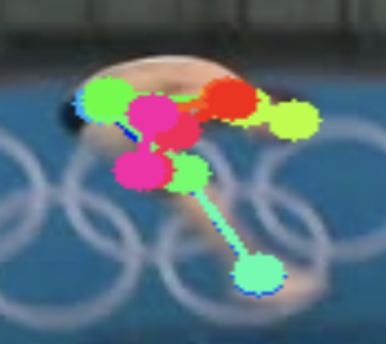}} \\ \bottomrule
\end{tabular}
}
\caption{\textbf{Symbols, their descriptions and visualizations}.}
\label{tab:symbols}
\end{table}

\subsection{Neural Action-Context Parser \raisebox{-0.7ex}{\includegraphics[scale=0.05]{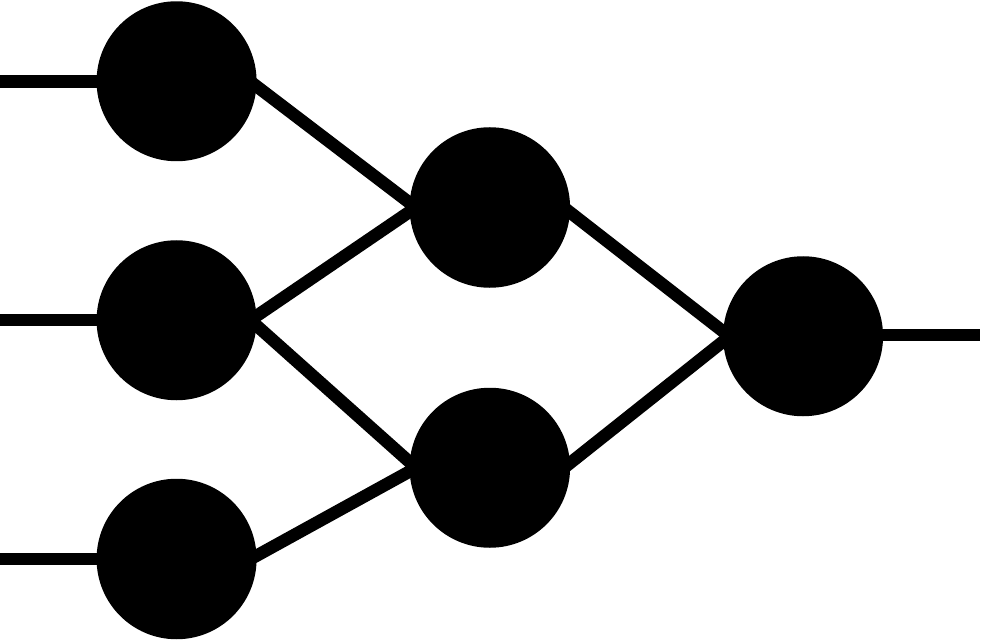}}} 
We implement the Neural Action-Context Parser as a collection of a dedicated: 1) \textit{splash detector}; 2) \textit{platform detector}, and 3) \textit{human pose estimator}. These object detectors \cite{wu2019detectron2} and pose estimators \cite{sun2019deep} first extract relevant information from each frame individually. For instance, we run a splash detector on all frames; whenever a splash is detected, it is segmented out. Object segmentation provides access to its properties, such as 1) \textit{size}, 2) \textit{shape}, and 3) \textit{position} within the scene. In cases where no splash is detected, all properties are filled with null values. Similarly, pose estimators record \textit{2D positions} of all \textit{major joints} of the diver. Direct information from these detectors forms the \textbf{primary set of symbols}. We further \textit{derive relations} between the primary symbols, such as: 1) the \textit{angles made by the bones at joints}, and 2) the \textit{distance between the diver and the platform}. These form a \textbf{secondary set of symbols}. Additional information is provided in \autoref{tab:symbols}.

\subsection{Rules-based Action Analyzer \raisebox{-0.8ex}{\includegraphics[scale=0.05]{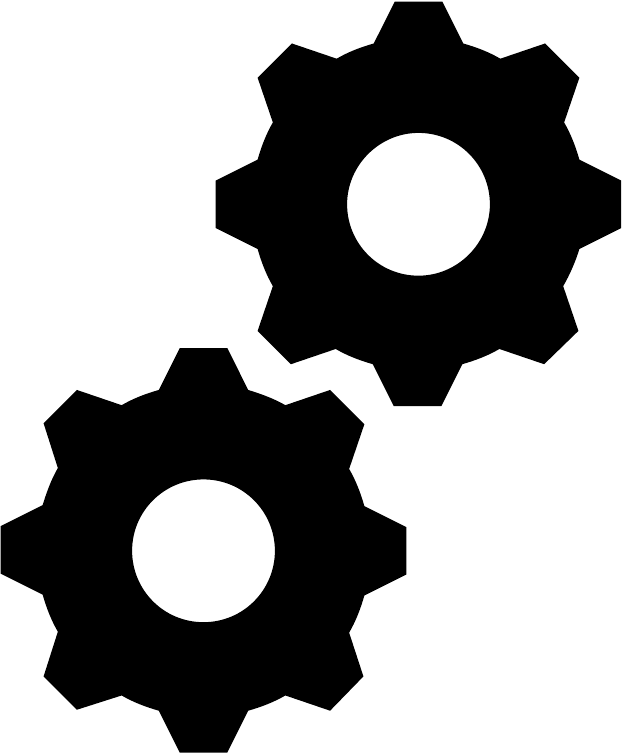}}}
This module takes in the symbols from the Neural Action-Context Parser and \textit{analytically} processes them using the rules formed by a domain expert\footnote{In our case, one of the authors is a diving expert with more than 10 years of experience.}. The rules are in accord with what is taught in the \textit{official USA Diving Judge's Course} \cite{Kwan}, and were further \textit{verified by 7 other domain experts}. In practice, these rules are implemented in \textit{Python} language \cite{van1995python} and termed as `\textbf{microprograms}.' Furthermore, action analysis is done in a \textit{hierarchical} manner. Specifically, the action analyzer: 1) first detects various \textit{class details of a dive}; 2) using these details, it then \textit{temporally segments the dive into its phases}; and 3) lastly, using the dive and its phase details, the \textit{quality of the dive is analyzed at a much finer granularity}. We discuss each level of this hierarchy and how they aid in the processing of the subsequent level in detail in the following.

\subsubsection{Detailed Dive Recognition} 
\paragraph{Preliminary.} In competitive diving, dives are identified on the basis of their: 1) \textbf{dive group} (\textit{forward, backward, reverse, inward, twisting, armstand}); 2) \textbf{number of somersaults} (number of somersaults are chosen in \textit{half-rotation increments}); 3) \textbf{number of twists} (chosen in \textit{half-rotation increments}); 4) \textbf{body position} (\textit{straight,  pike, tuck, free}). 

\begin{figure}
\small
    \centering
    \includegraphics[width=\columnwidth]{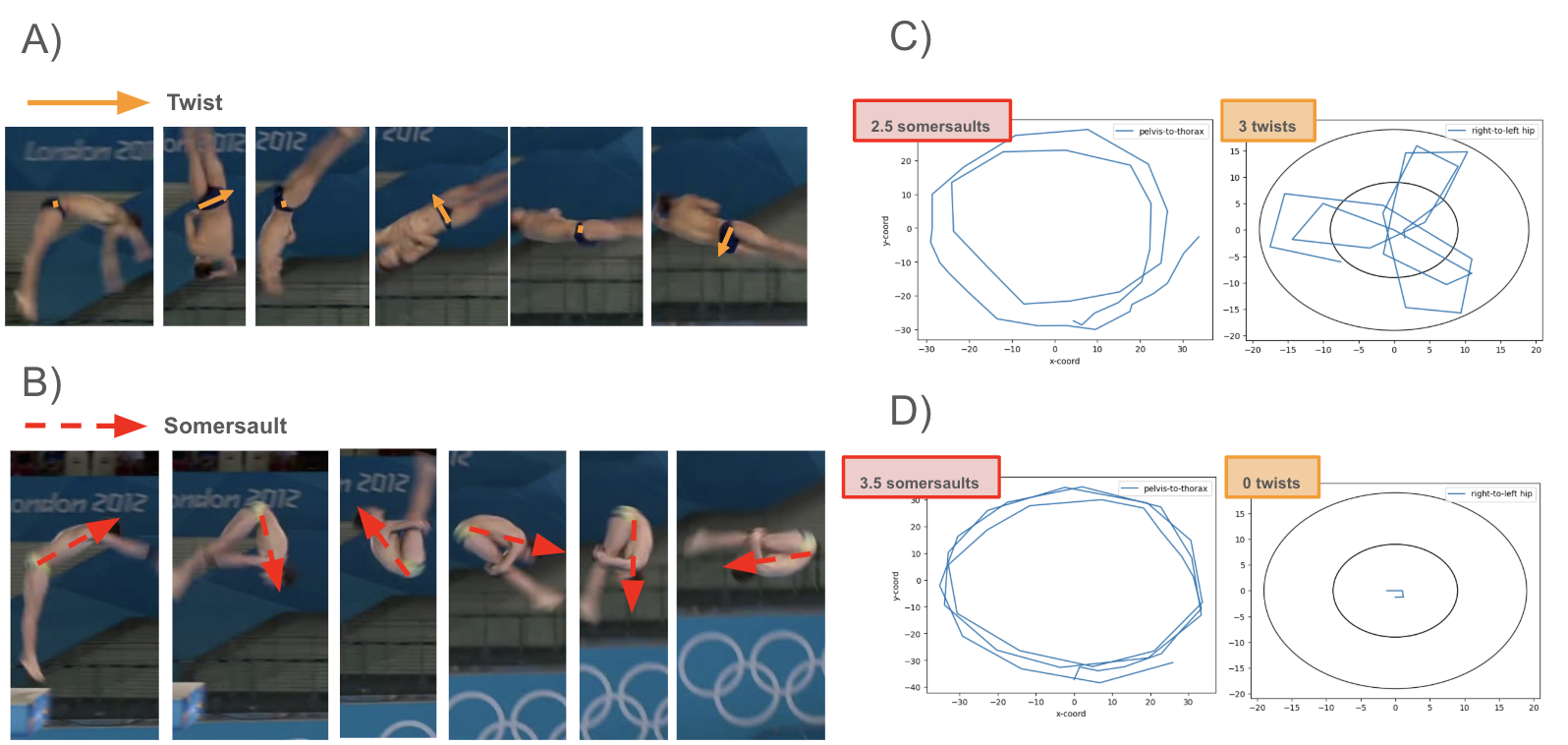}
    \caption{\textbf{Somersault and Twist Counters Visualized}. 
    \colorbox{lime}{\textbf{A)}}: Visualization of \textcolor{Orange}{right-to-left hip vector}. We count \textbf{twists} by counting the "\textbf{petals}" formed by this vector. 
    \colorbox{lime}{\textbf{B)}}: Visualization of \textcolor{red}{pelvis-to-thorax vector}. We count \textbf{somersaults} by \textit{counting the rotations} of this vector over the course of the dive. 
    \colorbox{lime}{\textbf{Graphical Solutions (C\&D)}}: We then show how the somersault and twist counters are applied to two different dives in C\&D. The \textcolor{blue}{blue trace} represents the vector we track, and the \textbf{black} circles in the twist plots represent the \textit{boundaries} of when "petals" are counted. The petal must surpass the inner black circle while staying inside the outer black circle to count as a petal. Each petal is 0.5 twists. 
    \colorbox{lime}{\textbf{C)}}: A forward dive with 2.5 somersaults (2.5 revolutions from the initial vector vertically up (\textbf{$\uparrow$}) to final vector vertically down ($\downarrow$) in the somersault plot) and 3 twists (6 “petals” in the twist plot). 
    \colorbox{lime}{\textbf{D)}}: A forward dive with 3.5 somersaults (3.5 revolutions from initial vector vertically up (\textbf{$\uparrow$}) to final vector vertically down ($\downarrow$) in the somersault plot) in pike position and no twists (zero “petals” in the twist plot).} 
    \label{fig:som_twist_counting}
\end{figure}

\paragraph{Rules-based dive recognizer.} Dive recognition with our neuro-symbolic approach\footnote{In contrast, recognizing a specific dive using a purely neural approach involves labeling a large number of dives to train a deep learning model.} is accomplished \textbf{analytically} from framewise pose data. Based on the abstracted symbols, the \textbf{dive group is identified} from: 1) \textit{the direction the diver is facing} and 2) \textit{the body movement} (\textit{clockwise} or \textit{counterclockwise} rotation) at the \textit{start of the dive}. The \textbf{body position} (straight, tuck, pike, or free) is detected by the \textit{degree of hip and knee bends}. \textbf{Somersault counts} are determined by \textit{tracking the rotation of the vector} that points from the diver's \textit{pelvis to thorax}. \textbf{Twists are counted} by \textit{tracking the magnitude and direction of the vector} that points from the \textit{diver's right to left hip joint}. For more details on how these rules are derived, see twisting and somersault vector plots of example dives in \autoref{fig:som_twist_counting}. In there, we have shown how we first obtain graphical solutions to, for example, somersault and twist counting. These graphical solutions are then implemented in Python language. The pseudocode for the somersault counting microprogram is provided in \textbf{Algorithm~\ref{alg:som_count}} for reference. Further details on the microprograms that execute dive recognition and additional implementation details are provided in the supplementary material.

\begin{algorithm}[t] 
\small
\caption{Somersault Counter Microprogram}
\label{alg:som_count}
\begin{algorithmic}
\State $half\_som\_count \gets 0$
\State $armstand \gets is\_armstand(dive\_frames)$
\State $curr\_pose \gets None$
\State $vertical\_up\_vector \gets [0,1]$
\State $vertical\_down\_vector \gets [0,-1]$
\For{each $frame$ in $dive\_frames$}
    \State $curr\_pose \gets get\_pose(frame)$
    \If{$curr\_pose$ is not None and $frame$ not in takeoff phase}
        \State $thorax \gets curr\_pose[thorax]$
        \State $pelvis \gets curr\_pose[pelvis]$
        \State $vector1 \gets thorax - pelvis$
        \If{($armstand$ and $half\_som\_count$ is odd) or (not $armstand$ and $half\_som\_count$ is even)}
            \State $vector2 \gets vertical\_down\_vector$
            \State $angle \gets get\_angle\_degrees(vector1, vector2)$ 
        \Else
            \State $vector2 \gets vertical\_up\_vector$
            \State $angle \gets get\_angle\_degrees(vector1, vector2)$
        \EndIf
        \If{$angle \leq 75$}
            \State {$half\_som\_count$ $\gets$ {$half\_som\_count + 1$}}
        \EndIf
    \EndIf
\EndFor
\State \Return {$half\_som\_count$}

\end{algorithmic}
\end{algorithm}

\subsubsection{Temporal Segmentation}
\begin{figure}[!t]
    \centering
    \includegraphics[width=\columnwidth]{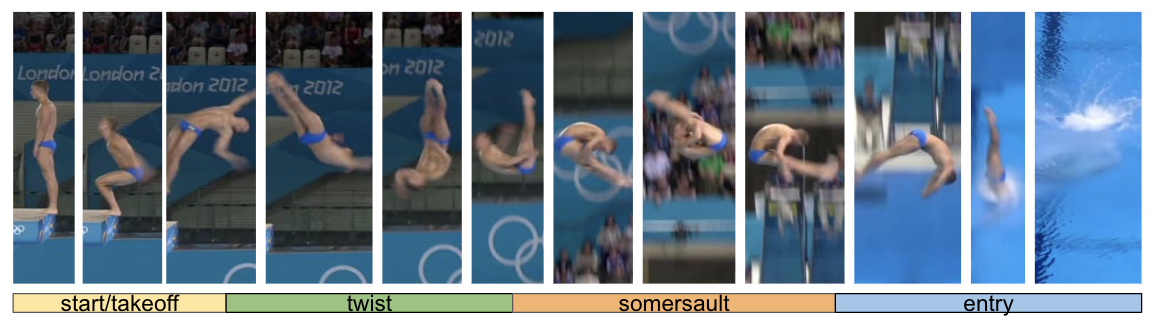}
    \caption{\textbf{Temporal Segmentation Visualization.} A dive is segmented into 4 \textit{phases}: \textit{start/takeoff}, \textit{twist}, \textit{somersault}, and \textit{entry}.} 
    \label{fig:temp_seg_visual}
\end{figure}

In this stage, the \textit{primary symbolic information}, along with \textit{dive class details} from the action recognition stage, is leveraged to perform the temporal segmentation of the dive. Each dive may be segmented into \textbf{start}/\textbf{takeoff}, \textbf{twist}, \textbf{somersault}, and \textbf{entry} phases (refer to \autoref{fig:temp_seg_visual}). Similar to detailed dive recognition, temporal segmentation is also accomplished \textbf{analytically} using \textbf{microprograms} that operate on \textit{symbols} and \textit{dive class details}---\textit{without the need to train a deep learning model with labeled video data} for this purpose. For example, to \textit{detect if a diver is in the start/takeoff phase}, we \textit{track the diver in relation to the end of the platform}. If the diver is \textit{beyond a threshold distance away from the platform}, this signifies the \textit{end} of the \textit{start/takeoff phase} in that the diver has \textit{jumped} and \textit{started} their dive. The \textit{other phases are detected} using similar \textit{logic-based algorithms}. Refer to the supplementary materials for specific implementation details.

\subsubsection{Fine-grained Dive (Action) Quality Assessment}
Now, we utilize the \textit{primary symbols} and output from the \textit{dive recognition and segmentation} to analyze them in extreme detail, as if  ``under a microscope." Similar to previous stages, action quality assessment is done \textbf{analytically} using our \textbf{AQA-micro-programs}. \textit{Output from previous stages} enables our \textbf{AQA-\textit{Meta}-program} to \textbf{select the correct microprogram} to execute to assess the quality of an element. For example, if the diver is detected to be in \textit{twisting phase}, then \textit{feet apart AQA microprogram} is executed to assess the performance w.r.t. feet apart aspect. As another example, \textit{meta-program} will \textit{not} execute \textit{Knee Straightness microprogram} if the diver is performing \textit{tuck dive}, because \textit{tuck position} does \textit{not} have \textit{Knee Straightness criteria}---an example is illustrated in last row of \autoref{tab:qualitative_comparison_to_neural}. We have discussed various elements of action/dive quality in \autoref{tab:elements_aqa_partial}. Further implementation details are provided in supplementary.

\begin{table}[]
\small
\centering
\resizebox{\columnwidth}{!}{%
\begin{tabular}{@{} p{0.9\columnwidth} c @{}}
\toprule
 \textbf{Distance-from-platform} of a diver is crucial for safety and aesthetic reasons. Divers risk injury or even death by hitting the concrete platform by being too close to it, leading to penalties. Conversely, being too far from the platform results in an unaesthetic trajectory and is also penalized.
 & \raisebox{-0.06\textwidth}{\includegraphics[width=0.09\textwidth, height=0.08\textwidth]{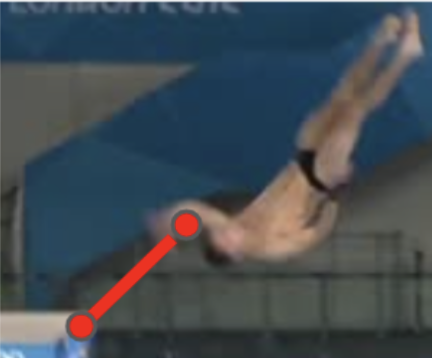}}\\ \midrule
 
 \textbf{Feet-apart.} A streamlined and graceful form throughout a dive is considered ideal.  Taking a cue from gymnastics, one aspect of an ideal diving form is maintaining one’s feet together.  A diver is penalized for having their feet apart. & \raisebox{-0.06\textwidth}{\includegraphics[width=0.09\textwidth, height=0.08\textwidth]{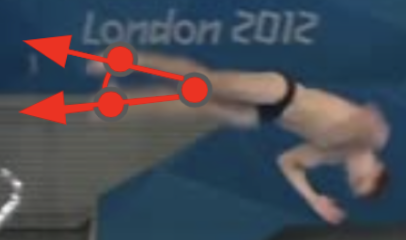}} \\ \midrule
 
 \textbf{Somersault tightness.} During a somersault, it is desirable for a diver to have as tight a tuck or pike position as possible.  A very tight somersault position allows a diver to spin faster and is considered ideal. 
 & \raisebox{-0.06\textwidth}{\includegraphics[width=0.09\textwidth, height=0.08\textwidth]{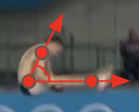}} \\ \midrule

 \textbf{Over/under-rotation.} Each particular dive has a specified number of rotations of somersaults to be performed.  The number of rotations is in half-rotation increments.  A diver preferably performs the exact number of rotations (to the degree) that are specified for the dive–no more and no less.  Over rotation and under rotation are penalized. 
 & \raisebox{-0.06\textwidth}{\includegraphics[width=0.09\textwidth, height=0.08\textwidth]{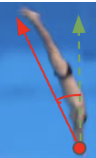}} \\ \midrule
 
 \textbf{Splash size.} Ideally, a diver has little or no splash upon entry into the water.  Little or no splash after a dive from 10 meters high is difficult to achieve and is indicative of very good form upon entry. 
 & \raisebox{-0.06\textwidth}{\includegraphics[width=0.09\textwidth, height=0.08\textwidth]{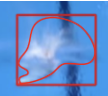}} \\
 
 \bottomrule
\end{tabular}
}
\caption{\textbf{Examples of elements of dive quality, their descriptions and visualizations.} Image annotations shown in \textcolor{red}{Red}. See all of the element descriptions/visualizations in supplementary.}
\label{tab:elements_aqa_partial}
\end{table}

\paragraph{Quantification of quality of each element.} 
We then quantify each of nine aspects of a dive as percentage scores, where 100\% is best (no error), and 0\% is worst (largest error). In other words, each error is given a score equal to its percentile ranking relative to all platform dives in the dataset. For example, for feet apart, if the average angle is smaller than 80\% of the dives, then the feet apart error would be scored 80\%. Advantageously, the \textit{percentile rankings inform the diver of their performance relative to their peers} which is the most relevant statistical information to have for a competitor trying to improve their results.  Furthermore, the scoring \textit{automatically adjusts to different levels of competition}. For example, scoring for high school divers would give percentiles relative to a dataset of high school divers. Note that, instead of percentiles, scores could be given on an absolute basis. For example, a fixed scale could convert the angle of feet apart to a score based on thresholds or a formula.  Advantageously, such a fixed scale would allow direct comparison across different levels of competition. However, such a fixed scale would be somewhat arbitrary and likely require value judgements to be made which are beyond the scope of this paper.  Hence, an absolute scoring system is not presented here. 

\paragraph{Overall Score.}
An overall score is obtained by \textbf{aggregating} the \textit{percentiles} of all the various aspects of the dive. We use \textit{uniform weighted averaging} as the aggregation function\footnote{According to the USA Diving Judge's Course \cite{Kwan}, each phase of the dive is supposed to be weighted equally. In particular, a judge can deduct up to 2 out of 10 points for each phase of the dive. Thus, we choose to average each aspect uniformly in order to give fairly equal importance to all of the aspects and phases of a dive.}. Other suitable functions may be used. By using \textit{percentiles}, we see how good the dive was compared to the level of dives in the semifinals and finals of Olympic and World Championship competitions. Using our system, it is also possible to use a weighted average instead of a uniformed average to emphasize certain aspects over others. While each phase is supposed to be weighted equally, errors that can occur within each phase have differing levels of importance depending on the personal preferences of the judges themselves. For example, some judges will weigh splash size more than verticalness, and some will do the exact opposite. Our system potentially allows for the \textbf{personalization} of weights for each error of the dive, giving each judge more flexibility in their scoring. Unlike conventional human judging, this personalization would be transparently presented, rather than hidden within a single overall score.

\paragraph{Visio-Linguistic Report generation.} 
Our NS-AQA system programmatically generates detailed performance reports in \textit{natural language}, based on the outputs discussed above. To do so, we take a simple approach without needing any large language models or deep learning models for language generation. We create a \textit{small library of template-sentences} with \textit{blanks to be filled in}. Again, using a \textit{Rules-based approach}, a \textit{microprogram} fetches an \textit{appropriate template-sentence} from the library. For example, for feet apart error, it would select to use the following template-sentence: {\small{\texttt{We found that your leg separation angle was on average \textcolor{red}{$<$insert average feet apart angle$>$} degrees for your dive}}}. The {\small{\texttt{\textcolor{red}{$<$insert average feet apart angle$>$}}}} is replaced with the average feet separation determined from the process discussed above. This process is repeated for all the errors. These errors and description sentences are \textit{compiled in the form of HTML page} as shown in \autoref{tab:detailed_reports}. Furthermore, our NS-AQA system also maintains a \textit{lookup-table} of error types with corresponding frame numbers and values of error severities. Using this lookup-table, it can \textit{retrieve} the frames contributing to a particular type of error and include them in the report as \textit{supporting visual feedback}. This detailed report is helpful for a number of reasons including as a support to human judges and as an educational tool to teach coaches, athletes, and judges how to score. More specific implementation details are provided in the supplementary.
\section{Results}
\label{sec:results}

We evaluate our system's performance on dive action recognition, temporal segmentation, and AQA. We compare temporal segmentation and action recognition performance to that of SOTA neural models. We evaluate AQA performance by surveying domain experts (Expert Survey) and comparing its output to a high performing purely neural model, C3D-MTL \cite{mtlaqa}.

\begin{table}
\small
\centering
\resizebox{\columnwidth}{!}{%
\begin{tabular}{@{}lcccc@{}} 
\toprule
\begin{tabular}[c]{@{}l@{}}\textbf{Action}\\ \textbf{Element}\end{tabular} 
& \multicolumn{1}{c}{\begin{tabular}[c]{@{}c@{}}\textbf{Nibali}\\ \textbf{\etal} \cite{nibali2017extraction}\end{tabular}} 
& \multicolumn{1}{c}{\begin{tabular}[c]{@{}c@{}}\textbf{MSCADC}\\ \cite{mtlaqa}\end{tabular}}
& \multicolumn{1}{c}{\begin{tabular}[c]{@{}c@{}}\textbf{C3D-MTL}\\ \cite{mtlaqa}\end{tabular}}
& \multicolumn{1}{c}{\begin{tabular}[c]{@{}c@{}}\textbf{Ours}\\ \textbf{NS-AQA}\end{tabular}}  \\
\midrule
Armstand & 98.30 & 97.45 &   99.72 &  \textbf{99.79} \\
Rotation Type & 78.75 & 84.70  &   97.45 &  \textbf{99.37} \\
Position & 74.79 & 78.47   &  96.32  &  \textbf{97.28} \\
No. of SS & 77.34 & 76.20  &   96.88 &  \textbf{97.31} \\
No. of TW & 79.89 & 82.72  &   93.20 &  \textbf{93.27} \\
\bottomrule
\end{tabular} 
}
\caption{\textbf{Performance evaluation on the task of fine-grained action recognition.} Accuracy (\%) is used as the performance metric. \textit{Higher is better.} SS: Somersaults; TW: Twists.} 
\label{tab:res_action_recognition} 
\end{table}

\subsection{NS Action Recognition}
We first evaluate the performance of our NS approach on detailed dive (action) recognition on MTL-AQA dataset \cite{mtlaqa}; \& compare its performance with SOTA methods \cite{mtlaqa}, which are purely neural in nature. \autoref{tab:res_action_recognition} shows that our NS action recognition approach outperforms C3D-MTL \cite{mtlaqa} on all categories. Note that, unlike neural approaches, \textit{our system also has the advantage of not requiring extra training specific to action recognition or temporal segmentation}.

\begin{table}
\small
\centering
\begin{tabular}{@{}lcc@{}} 
\toprule
\textbf{Model}   &  \textbf{AIoU@0.5} & \textbf{AIoU@0.75} \\
\midrule
TSA \cite{xu2022finediving}    &   82.51 &   34.31\\
Ours NS-AQA   &  \textbf{93.92}  &   \textbf{77.17} \\
\bottomrule
\end{tabular}
\caption{\textbf{Performance evaluation on the task of temporal segmentation.} Average Intersection over Union (AIoU) is used as the performance metric. \textit{Higher is better}.}
\label{tab:res_temporal_segmentation} 
\end{table}
\subsection{NS Action Temporal Segmentation}
Next, we evaluate the performance of our NS approach on temporal segmentation on FineDiving dataset \cite{xu2022finediving}, and compare it with SOTA Temporal Segmentation Attention (TSA) model \cite{xu2022finediving} in \autoref{tab:res_temporal_segmentation}. We observe that Our NS temporal segmentation model outperforms TSA model \cite{xu2022finediving}. Furthermore, the performance gap widens as the segmentation precision is increased from Average Intersection over Union (AIoU)@0.5 to 0.75. 

\begin{table} 
\small
\begin{center} 
\begin{tabular}{@{}lr@{}}
\toprule
\textbf{Element}  &  \textbf{Agreement (\%)}\\
\midrule
Overall Score & 92.5\\
Feet Apart &  96.0\\
Height off Board & 97.3\\
Distance from Board & 92.1 \\
Somersault Tightness &  94.5 \\
Knee Straightness & 100.0\\
Twist Tightness & 97.6 \\
Verticalness (over/under-rotation) & 90.5\\
Body Straightness During Entry & 94.5\\
Splash Size &  98.0\\
\bottomrule
\end{tabular} 
\caption{\textbf{Expert Survey Results} \textit{Higher is better}.} 
\label{tab:survey-results} 
\end{center} 
\end{table}
\subsection{NS Action Quality Assessment}
\begin{table*}
\small
\centering
\resizebox{\textwidth}{!}{
\begin{tabular}{@{}ccc@{}}
\includegraphics[width=0.33\textwidth, height=0.32\textwidth]{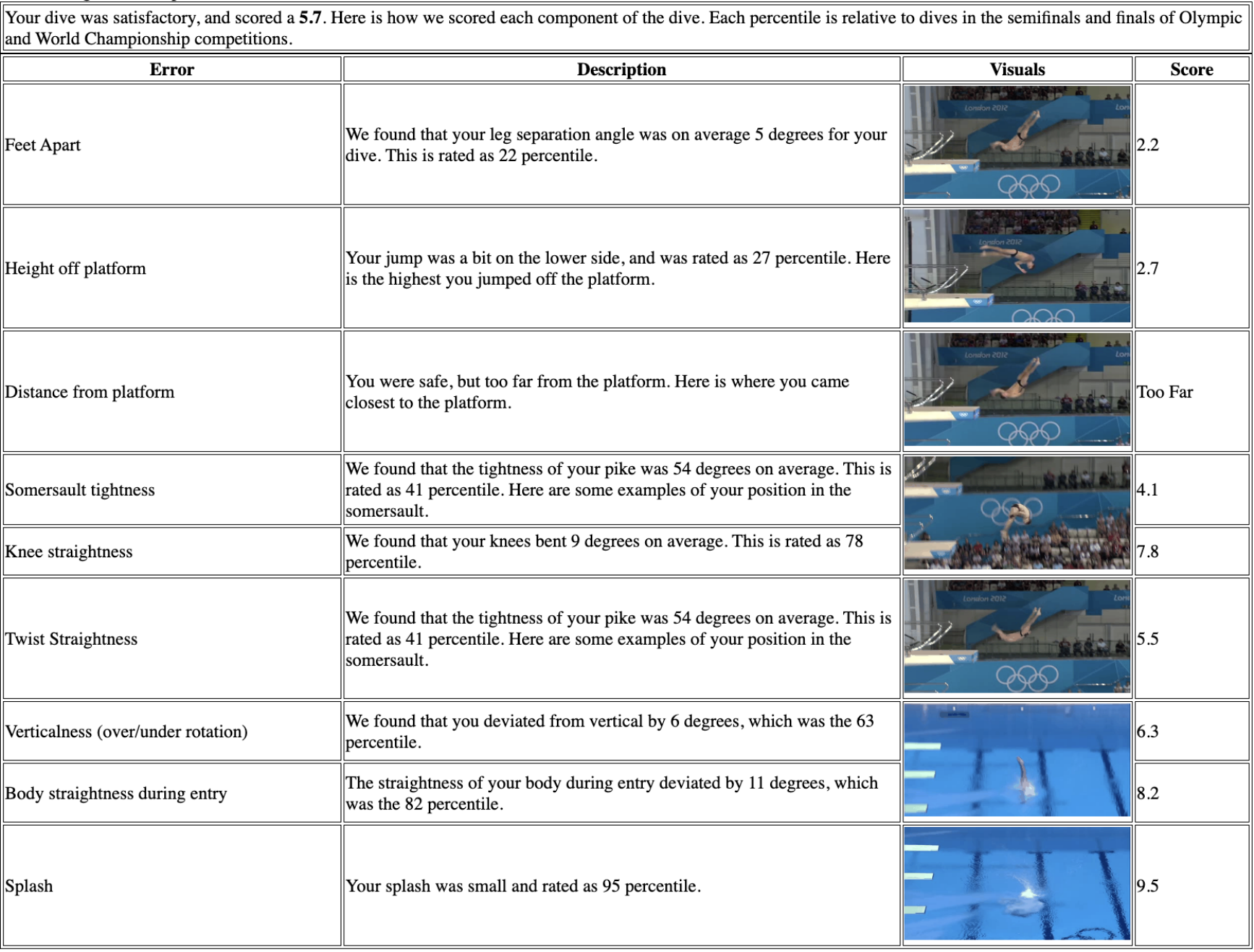} & \includegraphics[width=0.33\textwidth, height=0.32\textwidth]{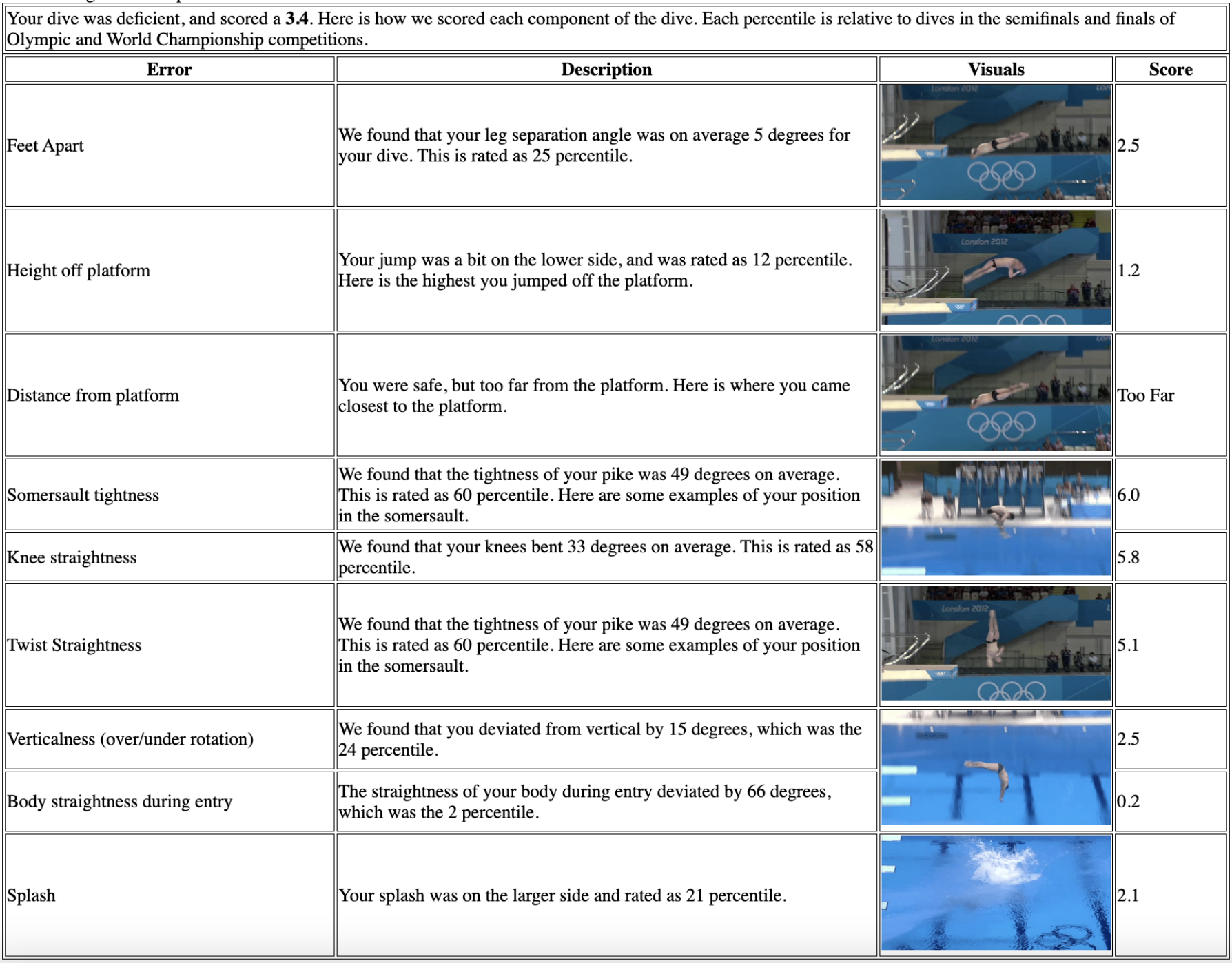} & \includegraphics[width=0.33\textwidth, height=0.32\textwidth]{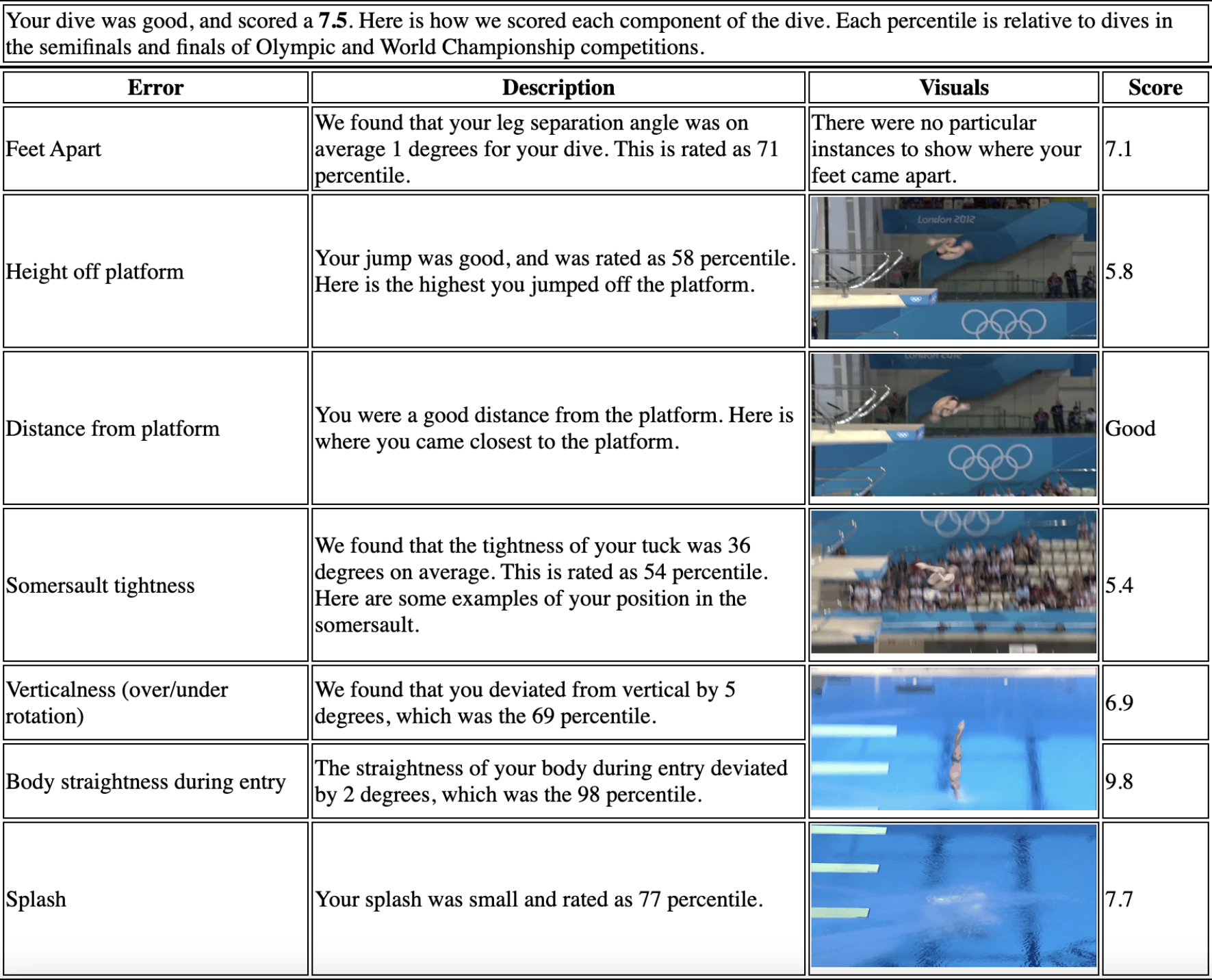} \\
\end{tabular} 
}
\caption{\textbf{Detailed and accurate reports (three samples shown here) programmatically generated by our NS-AQA approach.} \textit{Please zoom in}. More samples in supplementary.}
\label{tab:detailed_reports} 
\end{table*}

Unlike temporal action segmentation \& recognition, we do not evaluate our AQA scores on ``ground-truth" labels because our system proposes a new objective, and comprehensive way of scoring dives. Thus, our predicted scores are expected to be different than the subjective scores given by human judges of the current scoring system (as mentioned in Sec.~\ref{sec:introduction}). To evaluate model's performance in the absence of groundtruth labels, we ask a group of domain experts to evaluate the accuracy of our model \& give their feedback.

\subsubsection{Expert Survey}
We surveyed \textit{6 domain experts} to verify the accuracy of our system's outputted scores \& reports. These domain experts include trained judges, divers, \& coaches who have judged/competed/coached on the international \& national level. We showed each expert 50 randomly chosen dives from MTL-AQA dataset \cite{mtlaqa} \& asked whether they agreed or disagreed with each dive's overall score \& individual error scores. The results are shown in \autoref{tab:survey-results}. Each error (feet apart, distance from board, etc.) was approved by all experts as objective criteria that should be used to score the dive.  Experts agreed with our system on all categories at least 90\% of the time.

\subsubsection{Expert Feedback}
To validate the practical usefulness of our system in relation to diving competitions \& training, we presented our system \& its automatically generated reports (explained in Approach) to an \textit{Olympic diving coach} (referred to as Coach A for anonymity), \& \textit{a certified judge on the USA Diving Judges Commission} (referred to as Judge B). Both of them have generously offered valuable feedback for our system.

\paragraph{Expert Opinion 1: Coach A.}
According to Coach A, human judges emphasize overall impression when scoring a dive. This is because they cannot always see all of the diver’s mistakes in the short amount of time watching the dive so as to come up with a score based on specific errors. Even if judges try to be fair to all aspects and phases of a dive, as humans, it is almost impossible to see the dive perfectly at all times. Moreover, judging based on overall impression is inherently subjective and vulnerable to bias. Coach A believes that with the support of our NS-AQA system, scoring in diving can be ``\textit{more precise and have less errors.}" Coach A believes that this system, with its generated detailed report, would be helpful in a number of ways. In particular, it would be useful as a tool to (1) teach judges how to score, (2) catch any errors that a human judge may miss, (3) settle disagreements between judges, and (4) encourage safety by penalizing dangerous actions (e.g. getting too close to hitting the board) that lots of human judges overlook. See full opinion in supplementary.

\paragraph{Expert Opinion 2: Judge B.}
Judge B envisioned our system as an educational tool for teaching judges, coaches, and divers how to break down the dive into all of its components. ``\textit{We’re humans, we can’t get it all right,}” Judge B says, especially since judges only get 3 seconds to score a dive. Judge B particularly emphasized our system’s potential to improve diving safety due to its automated ``distance from board” measurement. In diving, hitting the platform can be the difference between life and death, so it is very important to prioritize the diver's safety. Judge B is ``\textit{very excited for this system to be able to make that call for us},” as it moves the \textit{villain}-role from the human judges to the AI to make those difficult decisions. See full opinion in supplementary.

\begin{table}
\small
    \centering
    \begin{tabular}{@{}lr@{}}
    \toprule
        \textbf{Model Output} & \textbf{Expert Preference} (\%)\\
        \midrule
        Fully Neural Score & 3.9 \\
        Our Neuro-Symbolic Report & \textbf{96.1} \\
        \bottomrule
    \end{tabular}
    \caption{\textbf{Comparison of experts' preference for Our NS model vs. fully neural models.} \textit{Higher is better.}}
    \label{tab:expert_pref}
\end{table}
\begin{table}
\small
\centering
\begin{tabular}{@{} c | c @{}}
\toprule
\textbf{Successes} & \textbf{Failures} \\
\midrule
    \begin{tabular}{c @{\hspace{.5em}} c}
        \includegraphics[width=.2\linewidth, height=0.15\linewidth]{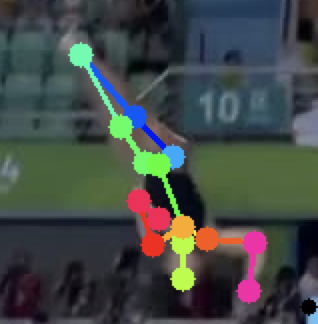} & 
        \includegraphics[width=.2\linewidth, height=.15\linewidth]{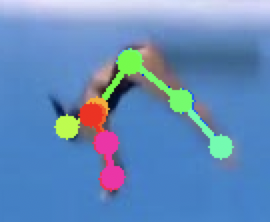} \\
        \includegraphics[width=.2\linewidth, height=.15\linewidth]{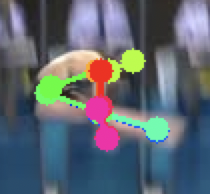} &
        \includegraphics[width=.2\linewidth, height=.15\linewidth]{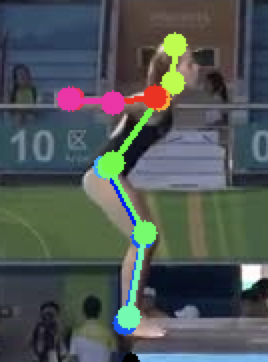}
    \end{tabular}
    &
    \begin{tabular}{c @{\hspace{.5em}} c}
        \includegraphics[width=.2\linewidth, height=0.15\linewidth]{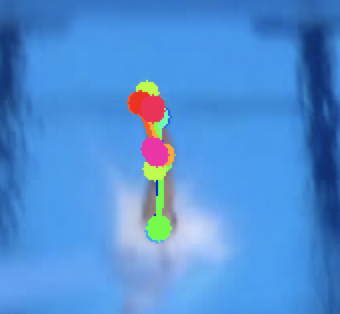} & 
        \includegraphics[width=.2\linewidth, height=.15\linewidth]{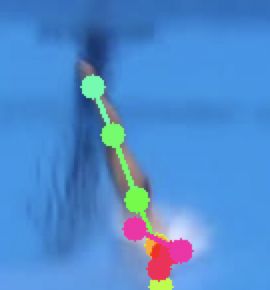} \\
        \includegraphics[width=.2\linewidth, height=.15\linewidth]{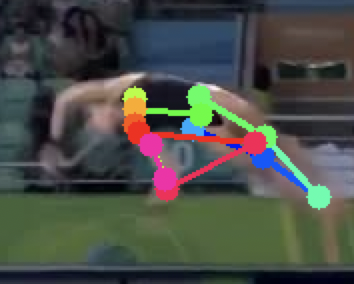} &
        \includegraphics[width=.2\linewidth, height=.15\linewidth]{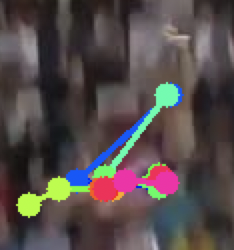}
    \end{tabular} \\
\bottomrule
\end{tabular}
\caption{\textbf{Pose estimation qualitative results.}}
\label{tab:pose_examples}
\end{table}
\begin{table*}[]
\small
\centering
\resizebox{0.97\textwidth}{!}{%
\begin{tabular}{@{}  c | p{0.27\textwidth} | p{0.17\textwidth} | p{0.35\textwidth} @{}}
\toprule
\textbf{Key Frames of Dive} & \textbf{Our Output} (\textit{Summarized, not full}) & \textbf{Neural Model Score} & \textbf{Comparison}\\
\midrule
   \begin{tabular}{c @{\hspace{.2em}} c}
    \includegraphics[width=.1\linewidth, height=.075\linewidth]{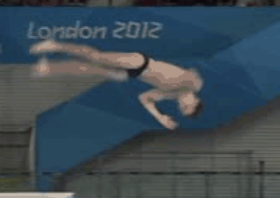} & 
    \includegraphics[width=.1\linewidth, height=.075\linewidth]{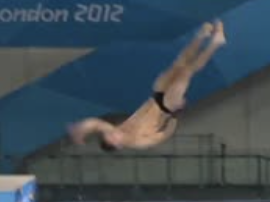} \\
    \includegraphics[width=.1\linewidth, height=.075\linewidth]{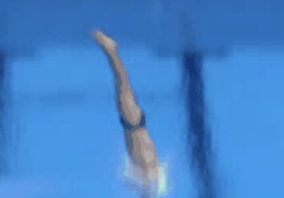} &
    \includegraphics[width=.1\linewidth, height=.075\linewidth]{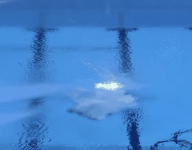}
  \end{tabular}  
 & 
  \begin{tabular}{l}
      \textbf{Overall Score: 5.7}\\
      Feet Apart: \textcolor{red}{2.2}\\
      Height off Platform: \textcolor{red}{2.7}\\
      Distance from Platform: \textcolor{red}{Too Far}\\
      Somersault Tightness: \textcolor{red}{4.1}\\
      Knee Straightness: \textcolor{ForestGreen}{7.8}\\
      Twist Straightness: \textcolor{ForestGreen}{5.5}\\
      Verticalness: \textcolor{ForestGreen}{6.3}\\
      Body Straightness during Entry: \textcolor{ForestGreen}{8.2}\\
      Splash: \textcolor{ForestGreen}{9.5}
  \end{tabular}  
 & 
 \vspace{-40pt}
 \textbf{Overall Score: 9.7} \textit{Note that, existing neural models only output a final score, unlike our model, which outputs a detailed report along with the final score.}
 & 
 \vspace{-55pt}
 The score outputted by C3D-MTL was significantly higher than our system. We see by looking at the dive's key frames that the \textbf{start/takeoff and flight was loose and sloppy}, but the entry was extremely good. This suggests that the \textbf{neural model was only factoring the entry} into the overall score instead of also accounting for the beginning parts of the dive. 
  \\ \midrule
  \begin{tabular}{c @{\hspace{.2em}} c}
    \includegraphics[width=.1\linewidth, height=0.075\linewidth]{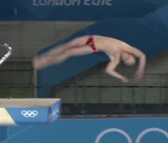} & 
    \includegraphics[width=.1\linewidth, height=.075\linewidth]{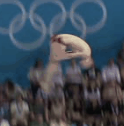} \\
    \includegraphics[width=.1\linewidth, height=.075\linewidth]{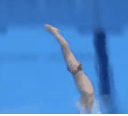} &
    \includegraphics[width=.1\linewidth, height=.075\linewidth]{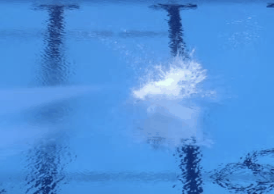}
  \end{tabular}  
 & 
  \begin{tabular}{l}
      \textbf{Overall Score: 5.4}\\
      Feet Apart: \textcolor{red}{3.9}\\
      Height off Platform: \textcolor{red}{0.1}\\
      Distance from Platform: \textcolor{ForestGreen}{Good}\\
      Somersault Tightness: \textcolor{red}{0.9}\\
      Knee Straightness: \textcolor{ForestGreen}{7.8}\\
      Twist Straightness: \textcolor{ForestGreen}{6.0}\\
      Verticalness: \textcolor{red}{4.8}\\
      Body Straightness during Entry: \textcolor{ForestGreen}{9.5}\\
      Splash: \textcolor{ForestGreen}{5.9}
  \end{tabular}  
 & 
 \textbf{Overall Score: 7.8}
 & 
 \vspace{-55pt}
 The score outputted by C3D-MTL was much higher than our system. We see by looking at the dive's key frames that the height off the platform was \textbf{very low} and the \textbf{pike in the somersault was not tight}. However, the entry was with \textbf{little splash}. This suggests that the \textbf{neural model was mostly factoring the splash} into the overall score instead of also accounting for factors like height and position tightness. \\ \midrule
 \begin{tabular}{c @{\hspace{.2em}} c}
    \includegraphics[width=.1\linewidth, height=0.075\linewidth]{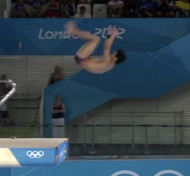} & 
    \includegraphics[width=.1\linewidth, height=.075\linewidth]{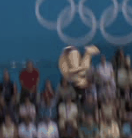} \\
    \includegraphics[width=.1\linewidth, height=.075\linewidth]{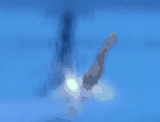} &
    \includegraphics[width=.1\linewidth, height=.075\linewidth]{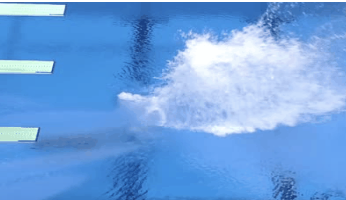}
 \end{tabular}  
 & 
  \begin{tabular}{l}
      \textbf{Overall Score: 6.4}\\
      Feet Apart: \textcolor{ForestGreen}{8.8}\\
      Height off Platform: \textcolor{ForestGreen}{8.5}\\
      Distance from Platform: \textcolor{ForestGreen}{Good}\\
      Somersault Tightness: \textcolor{ForestGreen}{6.0}\\
      Knee Straightness: N/A\\
      Twist Straightness: N/A\\
      Verticalness: \textcolor{red}{1.3}\\
      Body Straightness during Entry: \textcolor{ForestGreen}{9.8}\\
      Splash: \textcolor{red}{0.0}
  \end{tabular}  
 & 
 \textbf{Overall Score: 3.4}
 & 
 \vspace{-55pt}
 The score outputted by C3D-MTL was much lower than our system. Every aspect of the dive including the \textbf{height and  position tightness was very good} except for a slight mistiming when the diver came out of the dive, resulting in \textbf{over-rotation}. The diver should be rewarded for the other components of the dive they did well, instead of only penalized for their poor entry. \\ 
  
 \bottomrule
\end{tabular}}
\caption{\textbf{Qualitative comparison of our Neuro-Symbolic system to the purely Neural model C3D-MTL.} As taught in the official USA Diving Judge's Course, ``Dives should be judged as a whole, not emphasizing any single element. This approach in diving is especially true when it comes to the entry. It is easy to forgive earlier flaws if the dive has a good entry." \cite{Kwan} In all three examples, we see that the neural model emphasizes the entry of the dive far more than any other aspect, while our system more fairly accounts for each aspect.}
\label{tab:qualitative_comparison_to_neural}
\end{table*}
\subsubsection{Comparison to Neural Approach C3D-MTL}
We additionally surveyed domain experts to compare the predicted scores outputted by our system and neural models. We particularly compare our output to that of C3D-MTL. Experts were each shown 50 randomly chosen dives from MTL-AQA dataset \cite{mtlaqa}, and were asked to indicate which model output they agreed with more (either our NS model or the C3D-MTL model). We found that experts chose our system \textbf{96.1\%} of the time\footnote{For most of the dives when experts chose the neural output over ours, our system had failed pose estimation on some of the dive's frames. This affected the final score that was outputted.} (\autoref{tab:expert_pref}). A qualitative comparison of C3D-MTL and our system can be seen in \autoref{tab:qualitative_comparison_to_neural}.

\subsubsection{Failure Modes of Our System}
While our results are promising, our system occasionally fails to output a score that experts agree with. A main contributing factor to this is when the pose estimation on the diver is inaccurate. Particularly, pose estimation struggles when the diver is halfway underwater, and when the motion blur on the image is very strong. Examples of failed pose estimation compared with good pose are shown in \autoref{tab:pose_examples}. We believe that as pose estimation technology progresses, so will the accuracy of our system.
\section{Discussion}
We propose a neuro-symbolic (NS) paradigm for AQA, \& implement an NS-AQA system specifically for scoring competitive diving. Our NS system achieves SOTA results for fine-grained action recognition \& temporal segmentation, \& automatically generates a detailed \& comprehensive report scoring the performance of the dive. Our system was validated by domain experts including an Olympic diving coach \& a certified judge on the Judges Commission. Our approach ensures fairness and transparency, minimizing biases that often plague traditional assessment methods. Unlike traditional computer vision-based systems \& human judges that offer single scores without explanation, our NS approach provides a breakdown of assessments, empowering athletes \& coaches with actionable insights.

\noindent\textbf{Extension to other actions and skills assessment.} Although this paper is geared towards diving, our NS approach is general in nature, readily extensible to other sports like the vault in gymnastics \& figure skating. Domain expertise would be required to formulate rules. Our approach can be extended to other critical domains like surgery skills assessment as well. In surgery skills assessment, for example, microprograms to detect bleeding (analogous to splash in our case), respect for tissue, etc. could be developed. Final rating would be an aggregation of all such factors. We hope to inspire this work's application across diverse domains.

{{\noindent\textbf{Acknowledgement.} LO was supported by Princeton University.}}

{
    \small
    \bibliographystyle{ieeenat_fullname}
    \bibliography{main}

\begin{thebibliography}{33}
\providecommand{\natexlab}[1]{#1}
\providecommand{\url}[1]{\texttt{#1}}
\expandafter\ifx\csname urlstyle\endcsname\relax
  \providecommand{\doi}[1]{doi: #1}\else
  \providecommand{\doi}{doi: \begingroup \urlstyle{rm}\Url}\fi

\bibitem[Capecci et~al.(2019)Capecci, Ceravolo, Ferracuti, Iarlori, Monteriu, Romeo, and Verdini]{capecci2019kimore}
Marianna Capecci, Maria~Gabriella Ceravolo, Francesco Ferracuti, Sabrina Iarlori, Andrea Monteriu, Luca Romeo, and Federica Verdini.
\newblock The kimore dataset: Kinematic assessment of movement and clinical scores for remote monitoring of physical rehabilitation.
\newblock \emph{IEEE Transactions on Neural Systems and Rehabilitation Engineering}, 27\penalty0 (7):\penalty0 1436--1448, 2019.

\bibitem[Dadashzadeh et~al.(2024)Dadashzadeh, Duan, Whone, and Mirmehdi]{dadashzadeh2024pecop}
Amirhossein Dadashzadeh, Shuchao Duan, Alan Whone, and Majid Mirmehdi.
\newblock Pecop: Parameter efficient continual pretraining for action quality assessment.
\newblock In \emph{Proceedings of the IEEE/CVF Winter Conference on Applications of Computer Vision}, pages 42--52, 2024.

\bibitem[d'Avila Garcez et~al.(2012)d'Avila Garcez, Broda, and Gabbay]{Garcez2012NeuralsymbolicLS}
Artur~S. d'Avila Garcez, Krysia Broda, and Dov~M. Gabbay.
\newblock Neural-symbolic learning systems - foundations and applications.
\newblock In \emph{Perspectives in Neural Computing}, 2012.

\bibitem[Doughty et~al.()Doughty, Damen, and Mayol-Cuevas]{doughty2017s}
Hazel Doughty, Dima Damen, and Walterio Mayol-Cuevas.
\newblock Who’s better, who’s best: Skill determination in video using deep ranking.

\bibitem[Doughty et~al.(2018)Doughty, Damen, and Mayol-Cuevas]{doughty2018s}
Hazel Doughty, Dima Damen, and Walterio Mayol-Cuevas.
\newblock Who's better? who's best? pairwise deep ranking for skill determination.
\newblock In \emph{Proceedings of the IEEE conference on computer vision and pattern recognition}, pages 6057--6066, 2018.

\bibitem[Doughty et~al.(2019)Doughty, Mayol-Cuevas, and Damen]{doughty2019pros}
Hazel Doughty, Walterio Mayol-Cuevas, and Dima Damen.
\newblock The pros and cons: Rank-aware temporal attention for skill determination in long videos.
\newblock In \emph{Proceedings of the IEEE/CVF conference on computer vision and pattern recognition}, pages 7862--7871, 2019.

\bibitem[Jain et~al.(2020)Jain, Harit, and Sharma]{jain2020action}
Hiteshi Jain, Gaurav Harit, and Avinash Sharma.
\newblock Action quality assessment using siamese network-based deep metric learning.
\newblock \emph{IEEE Transactions on Circuits and Systems for Video Technology}, 31\penalty0 (6):\penalty0 2260--2273, 2020.

\bibitem[Kwan()]{Kwan}
Amy Kwan.
\newblock Usa diving elearning center.

\bibitem[Lei et~al.(2019)Lei, Du, Zhang, Ye, and Chen]{lei2019survey}
Qing Lei, Ji-Xiang Du, Hong-Bo Zhang, Shuang Ye, and Duan-Sheng Chen.
\newblock A survey of vision-based human action evaluation methods.
\newblock \emph{Sensors}, 19\penalty0 (19):\penalty0 4129, 2019.

\bibitem[Li et~al.(2018{\natexlab{a}})Li, Chai, and Chen]{li2018end}
Yongjun Li, Xiujuan Chai, and Xilin Chen.
\newblock End-to-end learning for action quality assessment.
\newblock In \emph{Pacific Rim Conference on Multimedia}, pages 125--134. Springer, 2018{\natexlab{a}}.

\bibitem[Li et~al.(2018{\natexlab{b}})Li, Chai, and Chen]{li2018scoringnet}
Yongjun Li, Xiujuan Chai, and Xilin Chen.
\newblock Scoringnet: Learning key fragment for action quality assessment with ranking loss in skilled sports.
\newblock In \emph{Asian Conference on Computer Vision}, pages 149--164. Springer, 2018{\natexlab{b}}.

\bibitem[Li et~al.(2019)Li, Huang, Cai, and Sato]{li2019manipulation}
Zhenqiang Li, Yifei Huang, Minjie Cai, and Yoichi Sato.
\newblock Manipulation-skill assessment from videos with spatial attention network.
\newblock In \emph{Proceedings of the IEEE/CVF international conference on computer vision workshops}, pages 0--0, 2019.

\bibitem[Liang et~al.(2023)Liang, Wang, Miao, and Yang]{liang2023logic}
Chen Liang, Wenguan Wang, Jiaxu Miao, and Yi Yang.
\newblock Logic-induced diagnostic reasoning for semi-supervised semantic segmentation.
\newblock In \emph{Proceedings of the IEEE/CVF International Conference on Computer Vision}, pages 16197--16208, 2023.

\bibitem[Liu et~al.(2021)Liu, Li, Jiang, Wang, Miao, Shan, and Li]{liu2021towards}
Daochang Liu, Qiyue Li, Tingting Jiang, Yizhou Wang, Rulin Miao, Fei Shan, and Ziyu Li.
\newblock Towards unified surgical skill assessment.
\newblock In \emph{Proceedings of the IEEE/CVF Conference on Computer Vision and Pattern Recognition}, pages 9522--9531, 2021.

\bibitem[Liu et~al.(2020)Liu, Liu, Huang, Qiao, Hu, Jiang, Zhang, Liu, and Guo]{liu2020fsd}
Shenglan Liu, Xiang Liu, Gao Huang, Hong Qiao, Lianyu Hu, Dong Jiang, Aibin Zhang, Yang Liu, and Ge Guo.
\newblock Fsd-10: A fine-grained classification dataset for figure skating.
\newblock \emph{Neurocomputing}, 413:\penalty0 360--367, 2020.

\bibitem[Nibali et~al.(2017)Nibali, He, Morgan, and Greenwood]{nibali2017extraction}
Aiden Nibali, Zhen He, Stuart Morgan, and Daniel Greenwood.
\newblock Extraction and classification of diving clips from continuous video footage.
\newblock In \emph{Proceedings of the IEEE conference on computer vision and pattern recognition workshops}, pages 38--48, 2017.

\bibitem[Pan et~al.(2019)Pan, Gao, and Zheng]{pan2019action}
Jia-Hui Pan, Jibin Gao, and Wei-Shi Zheng.
\newblock Action assessment by joint relation graphs.
\newblock In \emph{Proceedings of the IEEE/CVF international conference on computer vision}, pages 6331--6340, 2019.

\bibitem[Parmar and Morris(2019)]{aqa7}
Paritosh Parmar and Brendan Morris.
\newblock Action quality assessment across multiple actions.
\newblock In \emph{2019 IEEE winter conference on applications of computer vision (WACV)}, pages 1468--1476. IEEE, 2019.

\bibitem[Parmar and Morris(2016)]{parmar2016measuring}
Paritosh Parmar and Brendan~Tran Morris.
\newblock Measuring the quality of exercises.
\newblock In \emph{2016 38th Annual International Conference of the IEEE Engineering in Medicine and Biology Society (EMBC)}, pages 2241--2244. IEEE, 2016.

\bibitem[Parmar and Tran~Morris(2017)]{ltsoe}
Paritosh Parmar and Brendan Tran~Morris.
\newblock Learning to score olympic events.
\newblock In \emph{Proceedings of the IEEE conference on computer vision and pattern recognition workshops}, pages 20--28, 2017.

\bibitem[Parmar and Tran~Morris(2019)]{mtlaqa}
Paritosh Parmar and Brendan Tran~Morris.
\newblock What and how well you performed? a multitask learning approach to action quality assessment.
\newblock In \emph{Proceedings of the IEEE Conference on Computer Vision and Pattern Recognition}, pages 304--313, 2019.

\bibitem[Sardari et~al.(2020)Sardari, Paiement, Hannuna, and Mirmehdi]{sardari2020vi}
Faegheh Sardari, Adeline Paiement, Sion Hannuna, and Majid Mirmehdi.
\newblock Vi-net—view-invariant quality of human movement assessment.
\newblock \emph{Sensors}, 20\penalty0 (18):\penalty0 5258, 2020.

\bibitem[Sun et~al.(2022)Sun, Tjandrasuwita, Sehgal, Solar-Lezama, Chaudhuri, Yue, and Costilla-Reyes]{sun2022neurosymbolic}
Jennifer~J Sun, Megan Tjandrasuwita, Atharva Sehgal, Armando Solar-Lezama, Swarat Chaudhuri, Yisong Yue, and Omar Costilla-Reyes.
\newblock Neurosymbolic programming for science.
\newblock \emph{arXiv preprint arXiv:2210.05050}, 2022.

\bibitem[Sun et~al.(2019)Sun, Xiao, Liu, and Wang]{sun2019deep}
Ke Sun, Bin Xiao, Dong Liu, and Jingdong Wang.
\newblock Deep high-resolution representation learning for human pose estimation, 2019.

\bibitem[Tang et~al.(2020)Tang, Ni, Zhou, Zhang, Lu, Wu, and Zhou]{usdl}
Yansong Tang, Zanlin Ni, Jiahuan Zhou, Danyang Zhang, Jiwen Lu, Ying Wu, and Jie Zhou.
\newblock Uncertainty-aware score distribution learning for action quality assessment.
\newblock In \emph{Proceedings of the IEEE/CVF Conference on Computer Vision and Pattern Recognition (CVPR)}, 2020.

\bibitem[Van~Rossum and Drake~Jr(1995)]{van1995python}
Guido Van~Rossum and Fred~L Drake~Jr.
\newblock \emph{Python tutorial}.
\newblock Centrum voor Wiskunde en Informatica Amsterdam, The Netherlands, 1995.

\bibitem[Wu et~al.(2019)Wu, Kirillov, Massa, Lo, and Girshick]{wu2019detectron2}
Yuxin Wu, Alexander Kirillov, Francisco Massa, Wan-Yen Lo, and Ross Girshick.
\newblock Detectron2.
\newblock \url{https://github.com/facebookresearch/detectron2}, 2019.

\bibitem[Xu et~al.(2019)Xu, Fu, Zhang, Chen, Jiang, and Xue]{xu2019learning}
Chengming Xu, Yanwei Fu, Bing Zhang, Zitian Chen, Yu-Gang Jiang, and Xiangyang Xue.
\newblock Learning to score figure skating sport videos.
\newblock \emph{IEEE transactions on circuits and systems for video technology}, 30\penalty0 (12):\penalty0 4578--4590, 2019.

\bibitem[Xu et~al.(2022)Xu, Rao, Yu, Chen, Zhou, and Lu]{xu2022finediving}
Jinglin Xu, Yongming Rao, Xumin Yu, Guangyi Chen, Jie Zhou, and Jiwen Lu.
\newblock Finediving: A fine-grained dataset for procedure-aware action quality assessment.
\newblock In \emph{Proceedings of the IEEE/CVF conference on computer vision and pattern recognition}, pages 2949--2958, 2022.

\bibitem[Yi et~al.(2018)Yi, Wu, Gan, Torralba, Kohli, and Tenenbaum]{yi2018neural}
Kexin Yi, Jiajun Wu, Chuang Gan, Antonio Torralba, Pushmeet Kohli, and Josh Tenenbaum.
\newblock Neural-symbolic vqa: Disentangling reasoning from vision and language understanding.
\newblock \emph{Advances in neural information processing systems}, 31, 2018.

\bibitem[Yu et~al.(2021)Yu, Rao, Zhao, Lu, and Zhou]{yu2021group}
Xumin Yu, Yongming Rao, Wenliang Zhao, Jiwen Lu, and Jie Zhou.
\newblock Group-aware contrastive regression for action quality assessment.
\newblock In \emph{Proceedings of the IEEE/CVF international conference on computer vision}, pages 7919--7928, 2021.

\bibitem[Zhang et~al.(2024)Zhang, Bai, Chen, Chen, Lu, Wang, and Tang]{zhang2024narrative}
Shiyi Zhang, Sule Bai, Guangyi Chen, Lei Chen, Jiwen Lu, Junle Wang, and Yansong Tang.
\newblock Narrative action evaluation with prompt-guided multimodal interaction.
\newblock \emph{arXiv preprint arXiv:2404.14471}, 2024.

\bibitem[Zhong and Demiris(2024)]{zhong2024dancemvp}
Yun Zhong and Yiannis Demiris.
\newblock Dancemvp: Self-supervised learning for multi-task primitive-based dance performance assessment via transformer text prompting.
\newblock In \emph{Proceedings of the AAAI Conference on Artificial Intelligence}, pages 10270--10278, 2024.

\end{thebibliography}
}


\end{document}